\documentclass[runningheads]{llncs}

\usepackage{booktabs}
\usepackage{makecell}
\usepackage{threeparttable}
\usepackage{multirow}

\usepackage{eccv}

\usepackage{eccvabbrv}
\usepackage{graphicx}
\usepackage[accsupp]{axessibility}
\usepackage{hyperref}
\usepackage{orcidlink}
\usepackage{xspace}

\newcommand{\methodname}{Know3D\xspace}

\begin{document}

\title{Know3D: Prompting 3D Generation with Knowledge from Vision-Language Models}
\titlerunning{Know3D}

\author{
Wenyue Chen\inst{1,2}$^{*}$ \and
Wenjue Chen\inst{4,2}$^{*}$ \and
Peng Li\inst{3} \and
Qinghe Wang\inst{5} \and
Xu Jia\inst{5} \and
Heliang Zheng\inst{2} \and
Rongfei Jia\inst{2} \and
Yuan Liu\inst{3}$^{\dagger}$ \and
Ronggang Wang\inst{1}$^{\dagger}$
}
\authorrunning{W. Chen et al.}

\institute{
Guangdong Provincial Key Laboratory of Ultra High Definition Immersive Media Technology, Shenzhen Graduate School, Peking University, Shenzhen, China
\and
Math Magic, Beijing, China
\and
The Hong Kong University of Science and Technology, Hong Kong, China
\and
Peking University, Beijing, China
\and
Dalian University of Technology, Dalian, China
}
\maketitle

\begingroup
\renewcommand\thefootnote{}
\footnotetext{$^*$ Equal contribution.}
\footnotetext{$^\dagger$ Corresponding authors.}
\endgroup

\begin{abstract}
 Recent advancements in 3D generation have significantly enhanced the fidelity and geometric details of synthesized 3D assets. However, due to the inherent ambiguity of single-view input and the lack of robust global structural priors--caused by limited 3D training data--the unseen regions generated by existing models remain stochastic and difficult to control. This often results in geometries that are either physically implausible or misaligned with user intent. In this paper, we propose Know3D, a novel framework designed to incorporate rich knowledge from Multimodal Large Language Models (MLLMs) into 3D generation processes. By leveraging latent hidden-state injection, Know3D supports language-controllable generation of the back-view for 3D assets. We utilize a VLM-diffusion-based architecture: the Vision Language Model~(VLM) is used to provide high-level semantic understanding, while the diffusion model serves as a bridge, transferring semantic knowledge into the 3D generation model. Extensive experiments demonstrate that Know3D effectively bridges the gap between abstract textual instructions and the geometric reconstruction of invisible regions. By transforming the traditionally stochastic back-view hallucination into a semantically controllable process, Know3D offers a promising direction for highly plausible and user-friendly 3D generation in the future. Project page: \url{https://xishuxishu.github.io/Know3D.github.io/}.

\end{abstract}

\section{Introduction}

High-quality 3D assets are essential to modern workflows across gaming, film, and embodied AI. Because manual modeling remains prohibitively labor-intensive, automating 3D asset generation has become a critical challenge for the vision and graphics communities. Recently, 3D generative modeling~\cite{xiang2025trellis2,xiang2025trellis,zhang2024clay,chen2025dora,hunyuan3d2025hunyuan3d,lai2025lattice,li2025sparc3d,he2025sparseflex,chen2025ultra3d,chen20253dtopia,wu2025direct3d} has advanced at a rapid pace. These breakthroughs have significantly enhanced both the fine-grained geometric details and visual fidelity of 3D assets.

\begin{figure}[t]
  \centering
  \includegraphics[width=1\linewidth]{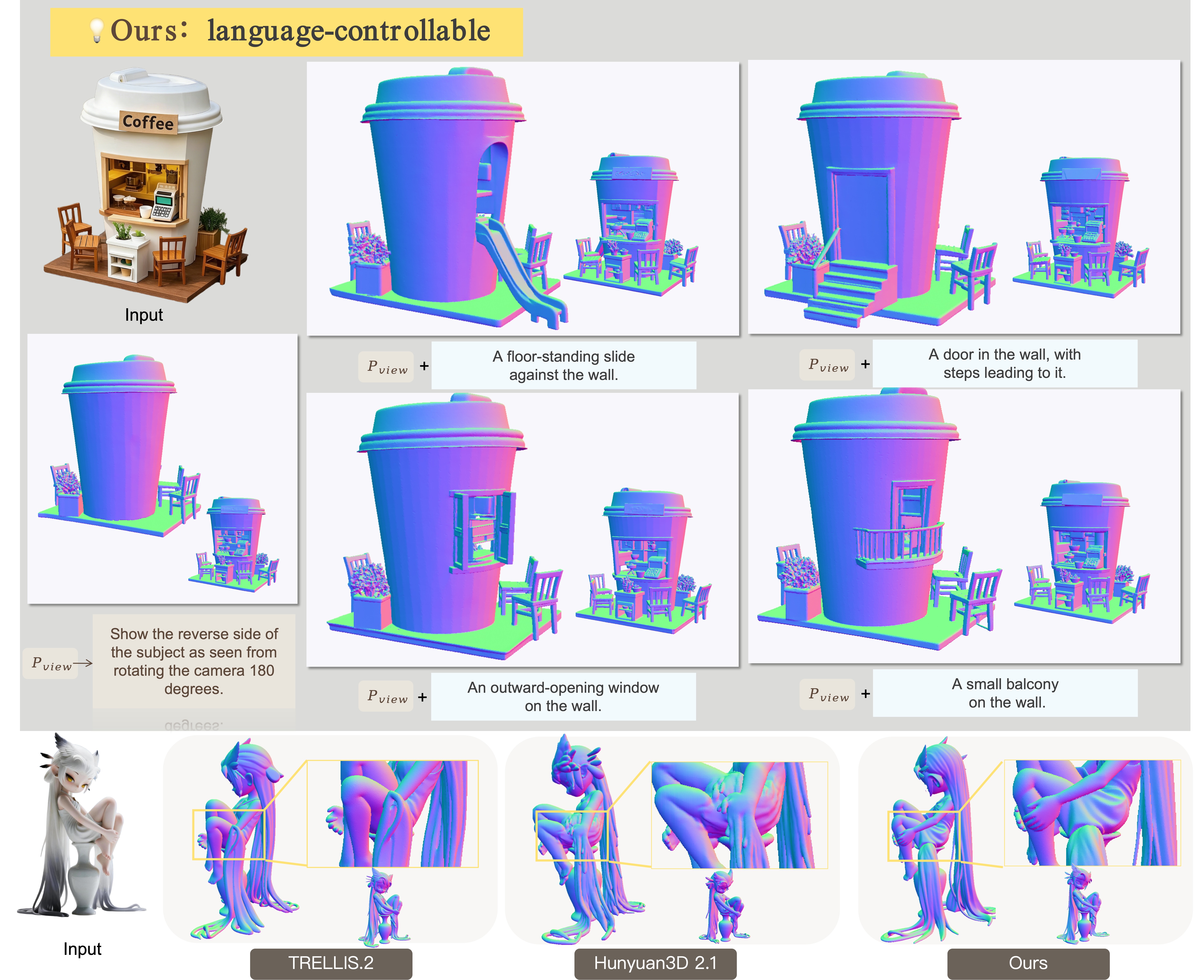}
  \caption{We propose Know3D, a knowledge-guided 3D generation framework that enables semantic control over the back-view of target objects. As shown in the coffee-cup scene, by simply varying the text, Know3D can synthesize diverse and structurally consistent back-view components. The bottom-row results demonstrate the potential of our method to improve the structural plausibility of unseen parts in 3D generation.}
  \label{fig:teaser} 
\end{figure}  

Despite these advancements, generating 3D assets from a single image remains a fundamentally ill-posed problem due to the inherent ambiguity of single-view observations. Image-to-3D generative models \cite{xiang2025trellis,xiang2025trellis2,hunyuan3d2025hunyuan3d} learn to map 2D observations to 3D shape by modeling the distribution of their training data. Since the input image only contains the information of the visible part, the synthesis of invisible parts relies on the model's internalized priors. While existing models are capable of hallucinating the occluded back-view, the synthesis remains predominantly stochastic and uncontrollable, and prone to two critical failure modes: (1) producing outputs that deviate from the user's creative or semantic intentions, as existing models inherently lack the ability to align the unobserved region synthesis with user-specified semantic constraints; (2) generating geometrically implausible structures that violate basic semantic commonsense constraints, as shown in Fig.~\ref{fig:teaser}. These failure modes can be largely attributed to data constraints. Compared to the internet-scale abundance of images, text, and videos, 3D datasets~\cite{deitke2023objaverse,objaverseXL,zhang2025texverse,hunyuan3d2026hy3d} are relatively limited in both quantity and diversity. Consequently, the world knowledge and structural common sense internalized by 3D generation models are naturally constrained. Further, high-quality, semantically aligned text-3D paired data remains less abundant.

Bridging this gap requires going beyond visual priors by incorporating additional knowledge, enabling models to infer unobserved structures from visible evidence. In particular, modern vision-language models (VLMs) have already learned rich semantic knowledge and commonsense reasoning from internet-scale multimodal data. If such knowledge can be effectively transferred to 3D generation models, it may provide valuable guidance for inferring unobserved object structures.

However, effectively prompting VLM knowledge into 3D generation presents its own challenges. A naive approach would be to directly use LLMs or VLMs to generate 3D representations in an autoregressive manner, as explored by recent shape LLM works~\cite{fang2025meshllm,ye2025shapellm,wang2024llama}. Yet, these methods have thus far underperformed compared to dedicated 3D generative models. Moreover, this autoregressive paradigm alters the model's pretrained knowledge priors. Forcing them into a constrained 3D generation task disrupts their original semantic capabilities. 

Another possible strategy is to directly feed VLM representations into 3D generation networks. However, such representations are typically highly abstract and lack explicit geometric grounding. As a result, they do not align well with the spatially structured feature spaces required for accurate 3D shape synthesis.

To address this challenge, we propose Know3D, a novel framework that prompts 3D generation with knowledge by leveraging the rich semantic understanding and commonsense reasoning capabilities of vision-language models (VLMs), thereby achieving enhanced controllability and better plausibility in 3D generation. Instead of directly injecting abstract
VLM representations, we leverage a multimodal diffusion model as
an intermediate bridge that translates semantic knowledge into
image-space structural priors.

Specifically, we utilize a VLM-diffusion-based model (Qwen-Image-Edit~\cite{wu2025qwen}), where the VLM~\cite{Qwen2.5-VL} is responsible for semantic understanding and provides guidance for image generation, and the diffusion model is responsible for generating images of the unobserved parts based on this guidance. The image-space structural priors serve as a medium to provide both semantic and structural information, thereby enabling semantically controllable 3D generation.

Although the Qwen-Image-Edit model demonstrates strong semantic understanding and can generate novel views based on prompts, it has two notable shortcomings: first, it often misinterprets spatial orientation, such as failing to accurately generate a ``back-view'' of an object; second, it frequently alters the subject's original pose or action in the output. Thus, we fine-tune Qwen-Image-Edit to improve the spatial awareness for better stability. Note that we annotate the corresponding textual description of the back-view for training to enable the generation control of the back-view.

We explored how to use the image-space structural priors as an intermediate medium to prompt the knowledge from VLMs into 3D generation models, experimenting with different designs for connecting them. Specifically, we experimented with (1) directly using the fully denoised VAE latent from MMDiT, (2) decoding this fully denoised VAE latent into an image and then extracting features via DINOv3~\cite{simeoni2025dinov3}, and (3) directly using the hidden states from the intermediate layers of MMDiT during the denoising process. Among these, directly using the hidden states from the intermediate layers of MMDiT demonstrates better spatial and semantic awareness and consequently achieves the best overall performance. 

Evaluations on HY3D-Bench~\cite{hunyuan3d2026hy3d} show that Know3D achieves competitive performance against state-of-the-art single-view 3D generation methods in semantic consistency with the conditional image. Moreover, our framework enables language-controllable generation of unseen backside regions, as shown in Fig.~\ref{fig:teaser}.

\section{Related Works}

\subsection{Native Single-view 3D Generation}
In recent years, native single-view 3D generation based on diffusion models has entered a period of rapid evolution, driven primarily by the advancement of 3D latent representations which have converged into two dominant paradigms: the Vector Set (VecSet)~\cite{zhang20233dshape2vecset,zhang2024clay,hunyuan3d2025hunyuan3d,li2025triposg,lai2025flashvdm,zhao2023michelangelo,jun2023shape,li2024craftsman3d,li2025step1x} approach that prioritizes global perception and high compression rates, and the Sparse Voxel approach~\cite{xiang2025trellis,xiang2025trellis2,li2025sparc3d,wu2025direct3d,ye2025hi3dgen,he2025sparseflex,ren2024xcube} that excels in local control and complex topological expression. Recent works have begun exploring the complementary fusion of these two paradigms. For instance, some approaches employ decoupled "coarse-to-fine" refinement frameworks to resolve the conflict between global structure and local geometry~\cite{lai2025lattice,chen2025ultra3d,jia2025ultrashape}, LATTICE~\cite{lai2025lattice} introduces semi-structured hybrid representations that inject spatial anchors into latent sets to enhance detail fidelity. Driven by these outstanding works, current models have achieved significant breakthroughs in both geometric and appearance fidelity. However, existing single-view 3D generation methods still have limitations in generating unobserved regions. This is mainly due to the limited information from a single view and the constraints of 3D training data.

\subsection{Text-to-3D Generation}
As a groundbreaking work in text-to-3D generation, DreamFusion~\cite{poole2022dreamfusion} pioneers score distillation from pre-trained 2D diffusion models to optimize 3D assets, with numerous subsequent works~\cite{poole2022dreamfusion,liang2024luciddreamer,lin2023magic3d,tang2023make,tang2023dreamgaussian,wang2023prolificdreamer} further refining this distillation pipeline for better generation performance. Native text-to-3D methods enable end-to-end generation directly in 3D representation spaces, with remarkable progress achieved in recent works~\cite {xiang2025trellis,zhao2025hunyuan3d,wu2025direct3d,li2025triposg,li2025step1x,zhao2023michelangelo,li2024craftsman3d}. Nevertheless, the controllability and fine-grained geometric accuracy of such paradigms still lag behind those of image-guided image-to-3D approaches. Some works~\cite{siddiqui2024meshgpt,chen2024meshanything,wang2024llama,ye2025shapellm,chen2025sar3d,fang2025meshllm,pun2025generating} explored Multimodal large language models for 3D generation,  yet these methods are constrained by limited representation resolution, failing to achieve high-fidelity 3D content generation.

\subsection{Unified Multimodal Models}
Recent research has focused on unified multimodal models for joint image understanding and generation. Existing methods fall into these paradigms: unified autoregressive models~\cite{team2024chameleon,wang2024emu3,wu2024vila,chen2025janus,geng2025x,sun2024generative,ge2024seed,tong2025metamorph}, unified diffusion models~\cite{li2025dual,shi2025muddit,swerdlow2025unified,wang2025fudoki,yang2025mmada}, decoupled LLM-diffusion frameworks~\cite{pan2025transfer,wu2025qwen,chen2025blip3,chen2025blip3o}, and hybrid AR-diffusion architectures~\cite{deng2025emerging,zhou2024transfusion,xie2024show}. While the 3D generation field has also been progressing rapidly in recent years, its overall development timeline lags behind that of the multimodal domain. Achieving semantic control in 3D generation remains challenging due to the limitation of 3D training data, which restricts the scale of multimodal pre-training compared to 2D domains. Furthermore, while text descriptions are highly abstract and lack geometric constraints, 3D generation requires explicit spatial, textural, and structural priors. To bridge this gap, we explore leveraging multimodal diffusion models as an intermediate bridge between vision-language knowledge and 3D generation. Instead of directly injecting abstract VLM representations, we observe that the intermediate hidden states of diffusion transformers encode rich spatial and structural information during the denoising process.

\begin{figure}[t]
  \centering
  \includegraphics[width=\linewidth]{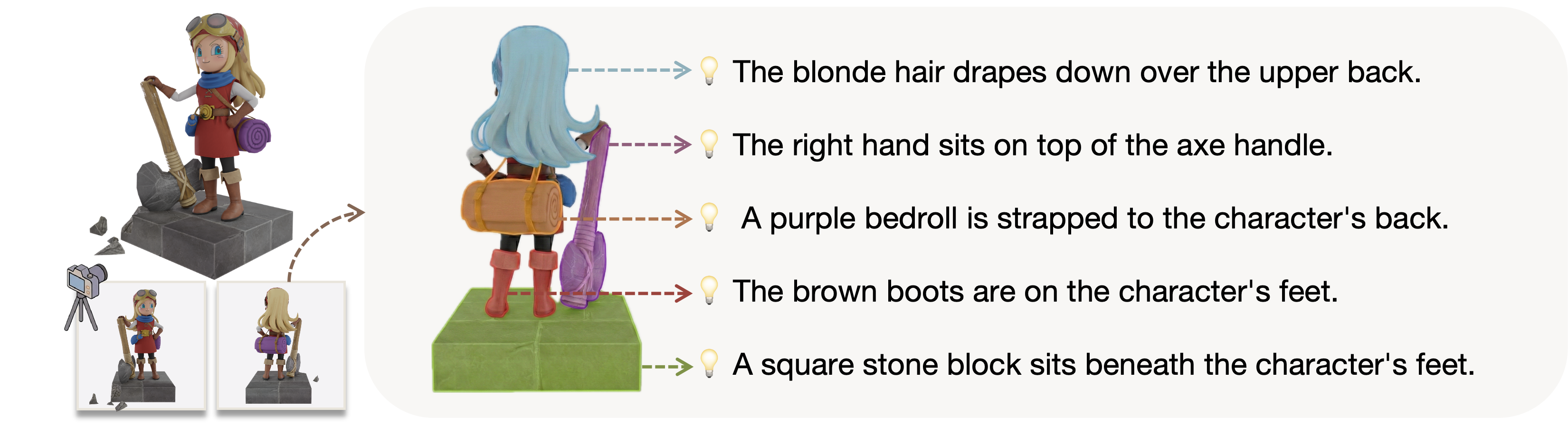}
  \caption{\textbf{Text annotation pipeline.} Given paired front and back views of a 3D asset, a VLM generates initial part-level descriptions.}
  \label{fig:qwen_image_training_data}
\end{figure}

\section{Method}
\begin{figure}[t]
  \centering
  \includegraphics[width=1\linewidth]{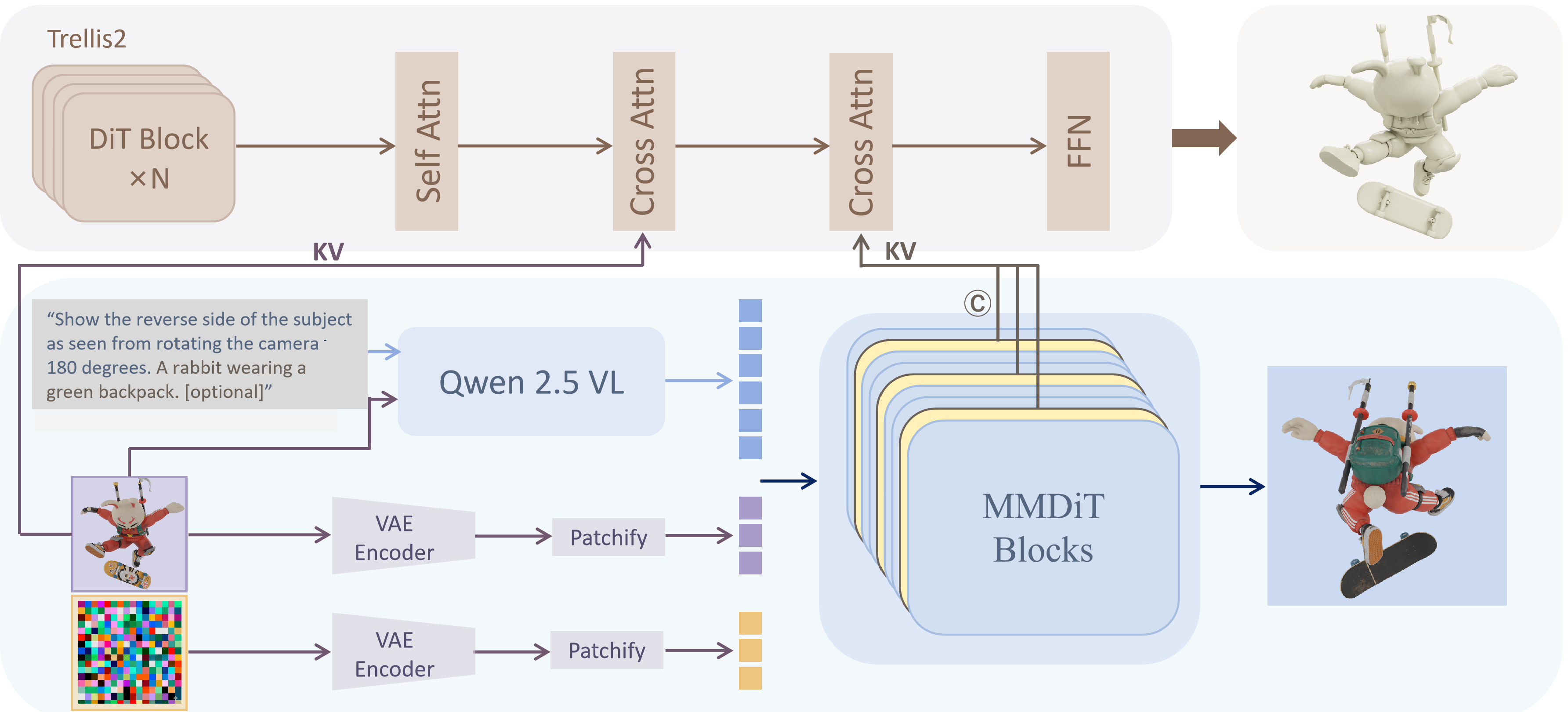}
  \caption{\textbf{Overview}. Given an input image and a text prompt, \methodname leverages Qwen-Image-Edit-2511~\cite{wu2025qwen}, which integrates Qwen2.5-VL with a Diffusion Transformer (DiT). Qwen2.5-VL~\cite{Qwen2.5-VL} encodes the multimodal inputs to guide a latent denoising process, enabling semantically aligned backside generation. The latent hidden states from intermediate DiT layers, which embed rich semantic and structural priors, are subsequently injected as conditioning signals into the 3D generation model to guide complete shape synthesis.}
  \label{fig:overall}
\end{figure}

\textbf{Overview.} Given a single input image \(I\) and a textual description \(T\) of the target object’s back-view, our goal is to synthesize a complete 3D representation \(\mathcal{V}\). To provide semantic cues for the unseen side, we fine-tune Qwen-Image-Edit on paired front-back view data to generate a plausible back-view image conditioned on both $I$ and $T$ (Section~\ref{subsec:semantic-aware-front-back-generation}). Then, we propose Know3D (Section~\ref{subsec:know3d-pipeline}), a knowledge-guided 3D generation framework that enables semantic control over the back-view of target objects. The overall pipeline is illustrated in Fig.~\ref{fig:overall}.

\subsection{Semantic-Aware Front-Back View Generation}
\label{subsec:semantic-aware-front-back-generation}

In this section, we aim to fine-tune Qwen-Image-Edit-2511~\cite{wu2025qwen} to generate reasonable back-view images from a given image. Even though the Qwen-Image-Edit model already shows a strong semantic understanding of the input image and can generate novel view images following the prompt. It still lacks a strong spatial awareness to understand the ``back-view'' and often generates images from incorrect viewpoints. In addition to viewpoint inaccuracies, it also tends to alter the subject's original pose during generation. Thus, we fine-tune Qwen-Image-Edit to improve the spatial awareness for better stability. Note that we annotate the corresponding textual description of the back-view for training to enable the generation control of the back-view.

\subsubsection{Dataset Construction}
We construct the training data from high-quality 3D assets.
For each asset, we render $2N$ images using uniform azimuth sampling with random elevation.
We select $N$ views as front views and pair each with its opposite view to form front--back pairs $(I_{\text{front}}, I_{\text{back}})$. To enable semantic control for back-view generation, we annotate textual descriptions for each front--back image pair. For each front--back pair \((I_{\text{front}}, I_{\text{back}})\), we annotate a set of textual descriptions for the salient components visible in the back-view. The output is a description set \(D = \{d_1, d_2, \dots, d_m\}\), where each \(d_i\) describes one back-view component, as shown in Fig.~\ref{fig:qwen_image_training_data}.

\subsubsection{Training Strategy and Objective.} To achieve stable back-view generation with text prompt control, we design a stochastic prompting strategy and optimize the model using the Conditional Flow Matching (CFM) objective~\cite{lipman2022flow,tong2023cfm}.

\paragraph{Stochastic Prompt Construction.} To enable controllable generation, we construct the conditioning prompt:
\begin{equation}
P = P_{\text{view}} \oplus P_{\text{back}},
\end{equation}
$P_{\text{view}}$ is a fixed prompt describing the $180^\circ$ camera rotation, 
while $P_{\text{back}}$ is randomly sampled from the component-level description set $D$ 
with probability $0.5$. This stochastic prompting strategy enables the model 
to learn both unconditional back-view generation and semantically controlled generation.

\paragraph{Training Objective.} Following Qwen-Image-Edit~\cite{wu2025qwen}, we extract multimodal hidden states \(H_{\text{vlm}} = \text{Qwen2.5-VL}(I_{\text{front}}, P)\) from the vision-language model~\cite{Qwen2.5-VL}, and obtain the spatial latent condition \(Z_{\text{front}} = \mathcal{E}(I_{\text{front}})\) via a VAE encoder~\cite{wan2025} \(\mathcal{E}\). Let \(Z_{back} = \mathcal{E}(I_{back})\) denote the target latent of the back-view. Given a noisy latent \(Z_t = (1-t)Z_{back} + t\epsilon\) at timestep \(t \in [0,1]\), the vector field estimator \(v_\phi\) is optimized to predict the velocity field via the Conditional Flow Matching objective~\cite{lipman2022flow,tong2023cfm}:
\begin{equation}
\mathcal{L}_{\text{CFM}}(\phi) = \mathbb{E}_{t, Z_{\text{back}}, \epsilon} \| v_\phi(Z_t, t, H_{\text{vlm}}, Z_{\text{front}}) - (\epsilon - Z_{\text{back}}) \|_2^2.
\end{equation}

\subsection{Prompting 3D Generation with VLMs}
\label{subsec:know3d-pipeline}

In this section, we introduce how to use image features as an intermediate medium to prompt knowledge from vision-language models (VLMs) for 3D generation models. The more straightforward approach is to directly inject the images generated by Qwen-Image. However, this process involves VAE decoding and the re-extraction of features by DINOv3~\cite{simeoni2025dinov3}, making the workflow relatively cumbersome. Moreover, it relies on high-precision pixel-level restoration. If the quality of the generated images is insufficient, erroneous results will directly impact the 3D generation process. An ideal feature should possess the following characteristics: (1) sufficient spatial awareness to facilitate learning by 3D generation models, and (2) a certain degree of semantic awareness and robustness. We found that the hidden states of the intermediate layers of MMDiT inherently possess strong spatial awareness and rich semantic information~\cite{huang2025much3d,li2026unraveling}, enabling them to better guide 3D generation. 

\subsubsection{Knowledge Extraction and Prompting.}
Qwen2.5-VL~\cite{Qwen2.5-VL} encodes the input front-view image and text prompt into high-level semantic features, while the VAE encoder extracts visual features from the front-view input. These representations guide the MMDiT through the full iterative denoising process. We extract intermediate latent hidden states from $n$ MMDiT layers at a specific denoising timestep $t$ \cite{tang2023emergent,huang2025much3d,baade2026latentforcing}, and concatenate these layer-wise features to form the structural-semantic conditioning signal \(H_{\text{DiT}} = \text{Concat}(h^{(1)}, h^{(2)}, \dots, h^{(n)})\).

Building upon TRELLIS2~\cite{xiang2025trellis2}, we design a parallel cross-attention branch for \(H_{\text{DiT}}\) injection. We retain the backbone’s original self-attention and image-conditioned cross-attention layers intact to avoid interference with pre-trained 3D generation priors.
\(H_{\text{DiT}}\) is first linearly projected and then layer-normalized to obtain the projected feature \(H'_{\text{DiT}}\), which serves as keys and values for the new cross-attention layer.
Its output is scaled by a zero-initialized linear layer for stable training. The modified residual fusion process is formulated as:
\begin{equation}
\label{eq:fusion}
\begin{aligned}
& F_{\text{sa}} = \text{Self-Attn}(F), \\
& \Delta F_{\text{img}} = \text{Cross-Attn}(F_{\text{sa}}, F_{\text{img}}, F_{\text{img}}), \\
& \Delta F_{\text{dit}} = \text{ZeroLinear}(\text{Cross-Attn}(F_{\text{sa}}, H'_{\text{DiT}}, H'_{\text{DiT}})), \\
& F_{\text{out}} = F + \Delta F_{\text{img}} + \Delta F_{\text{dit}}.
\end{aligned}
\end{equation}

\subsubsection{3D Geometry Generation}
With the structural-semantic signal \(H_{\text{DiT}}\) and front-view feature \(F_{\text{front}}\) as dual conditioning signals, our 3D generation follows the two-stage paradigm of TRELLIS2~\cite{xiang2025trellis2}:
\begin{equation}
\begin{gathered}
V_{\text{ss}} = D_{\text{ss}} \left( G_{\text{ss}} \left( z, \tau, F_{\text{front}}, H_{\text{DiT}} \right) \right), \\
V_{\text{geo}} = D_{\text{geo}} \left( G_{\text{geo}} \left( z, \tau, F_{\text{img}}, H_{\text{DiT}} \mid V_{\text{ss}} \right) \right),
\end{gathered}
\end{equation}
where \(z\) is standard Gaussian noise, \(\tau\) is the diffusion timestep. The first stage generates a coarse sparse structure \(V_{\text{ss}}\) to model the global topological prior, and the second stage recovers high-fidelity fine geometry \(V_{\text{geo}}\) conditioned on \(V_{\text{ss}}\).

\subsubsection{Training Objective.} Both stages are optimized with the Conditional Flow Matching (CFM) objective~\cite{lipman2022flow,tong2023cfm}, 
\begin{equation}
    \mathcal{L}_{3D} = \mathbb{E}_{\tau, x_0, \epsilon} \| v_\theta(x_\tau, \tau, F_{\text{img}}, H_{\text{DiT}}) - (\epsilon - x_0) \|_2^2,
\end{equation}
where \(x_0\) is the ground-truth 3D geometric latent, and \(\epsilon \sim \mathcal{N}(0, \mathbf{I})\) is standard Gaussian noise.

\section{Experiments}
\subsection{Experiment Setup}
\subsubsection{Dataset.} For Semantic-Aware Front-Back View Generation Training, we use 5k high-quality 3D assets selected from the TexVerse dataset~\cite{zhang2025texverse}. For each asset, the field of view (FoV) is sampled from $\{35^\circ, 50^\circ, 85^\circ, 105^\circ, 135^\circ\}$ and elevation from $[-15^\circ, 45^\circ]$, with both fixed per asset but randomized across assets. We render 12 uniformly spaced azimuthal views over $360^\circ$ per mesh, forming 6 front-back pairs, and annotate all of them. For 3D generation training, we use 60k meshes from TexVerse. For each mesh, we render two sets of views: a perturbation-free set and a perturbed set. In the perturbation-free set, azimuth views are spaced every 45° and grouped into four front–back pairs. In the perturbed set, we apply random FoV scaling and small angular offsets to simulate viewpoint perturbations, forming four perturbed pairs. For evaluation, we conduct quantitative analysis and comparisons with baselines on the HY3D-Bench~\cite{hunyuan3d2026hy3d} dataset. For the ablation study, we randomly selected a subset of 100 3D assets in TexVerse~\cite{zhang2025texverse} dataset that not in our training data.

\subsubsection{Training Details.}
In Semantic-Aware Front-Back View Generation, we adopt Qwen-Image-Edit-2511~\cite{wu2025qwen} as our foundational pre-trained model, and fine-tune it using Low-Rank Adaptation(LoRA)~\cite{hu2022lora}. All experiments are conducted on 32 NVIDIA A800 GPUs with a global batch size of 32. We set the rank of the LoRA adapter to 64 for all trainable attention layers of the backbone model. For 3D generation, we freeze the parameters of the pre-trained Qwen-Image-Edit-2511~\cite{wu2025qwen} to preserve its generalizable visual priors. For the original Trellis2 network, we apply LoRA~\cite{hu2022lora} fine-tuning with a rank of 64. In contrast, the newly added condition layers, which are designed to integrate semantic-aware front-back view signals, are fully fine-tuned to ensure effective modulation of the 3D generation process. This stage is trained on 32 NVIDIA A800 GPUs with a global batch size of 64.

\begin{figure}[!htbp]
\centering\includegraphics[width=1\linewidth]{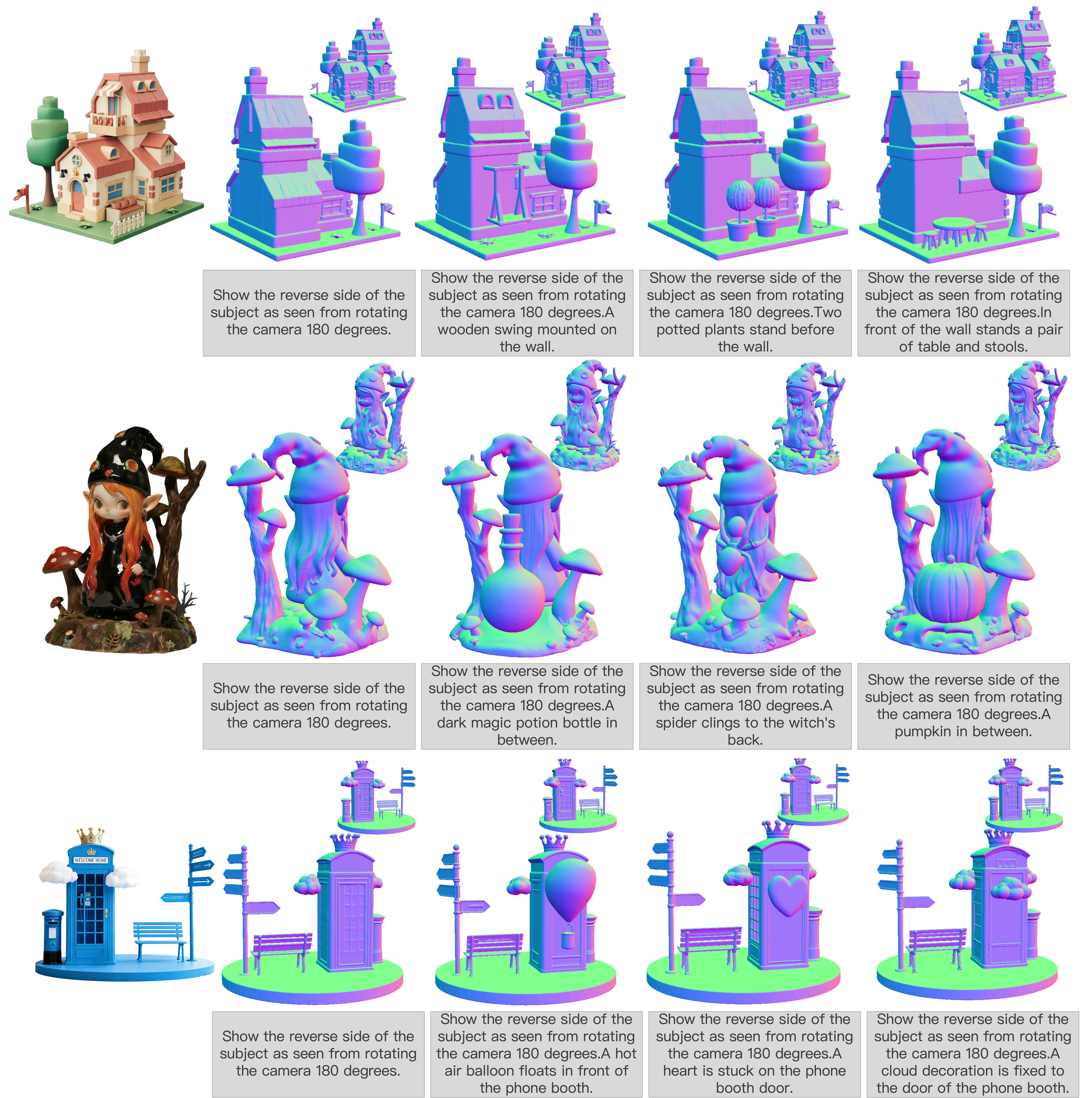}
  \caption{\textbf{Back-View Semantic Control Visualization.} The leftmost column shows the input front-view condition images. The remaining columns present back-view normal renderings of meshes generated under different text prompts, demonstrating flexible semantic control over back-side components.}
  \label{fig:senamic} 
\end{figure}  

\subsubsection{Metrics.}
To compare with baseline, we use ULIP~\cite{xue2023ulip} and Uni3D~\cite{zhou2023uni3d} to measure the semantic consistency between images and generated meshes. For the ablation study, we only trained the first stage (sparse voxel generation). Therefore, we evaluate the performance using IoU and Chamfer Distance (CD).

\begin{figure}[t] 
  \centering
  \includegraphics[width=0.9\linewidth]{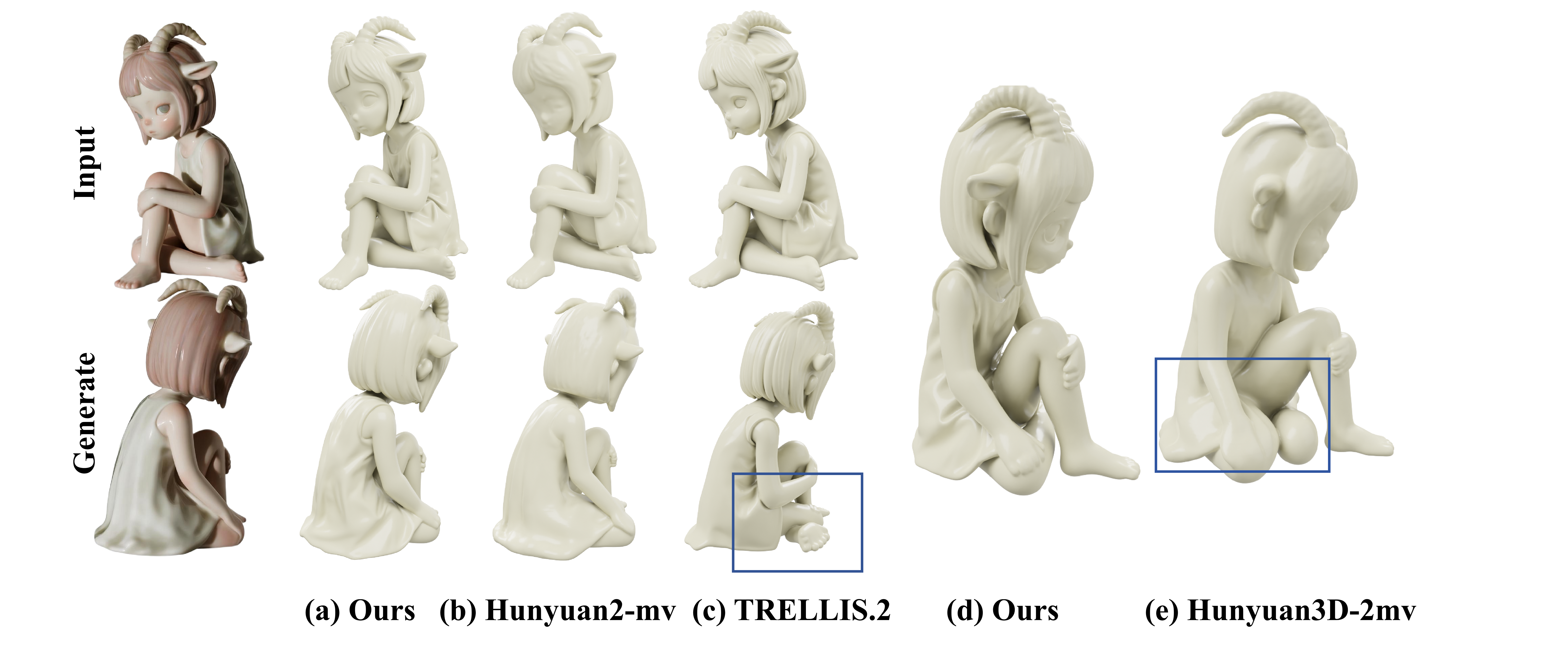} 
  \caption{\textbf{Visual Comparison between Know3D and Hunyuan3D-2mv.} As shown in (a) and (b), both Know3D and Hunyuan3D-2mv~\cite{hunyuan3d22025tencent} achieve reasonably good alignment between the input and generated viewpoints. However, for novel viewpoints, Hunyuan3D-2mv produces implausible structures(d), whereas Know3D still yields reasonable results(e).} 
  \label{fig:ab_mv} 
\end{figure}
\subsection{Generation Controllability and Quality}

In this section, we present a comparison of Know3D with current state-of-the-art methods, as well as a qualitative analysis of its semantic controllability.
\subsubsection{Comparison with Baselines.}
We evaluate the generation quality of Know3D by comparing with single- and multi-view 3D generation methods.

\begin{figure}[!htbp] 
  \centering
\includegraphics[width=0.9\linewidth]{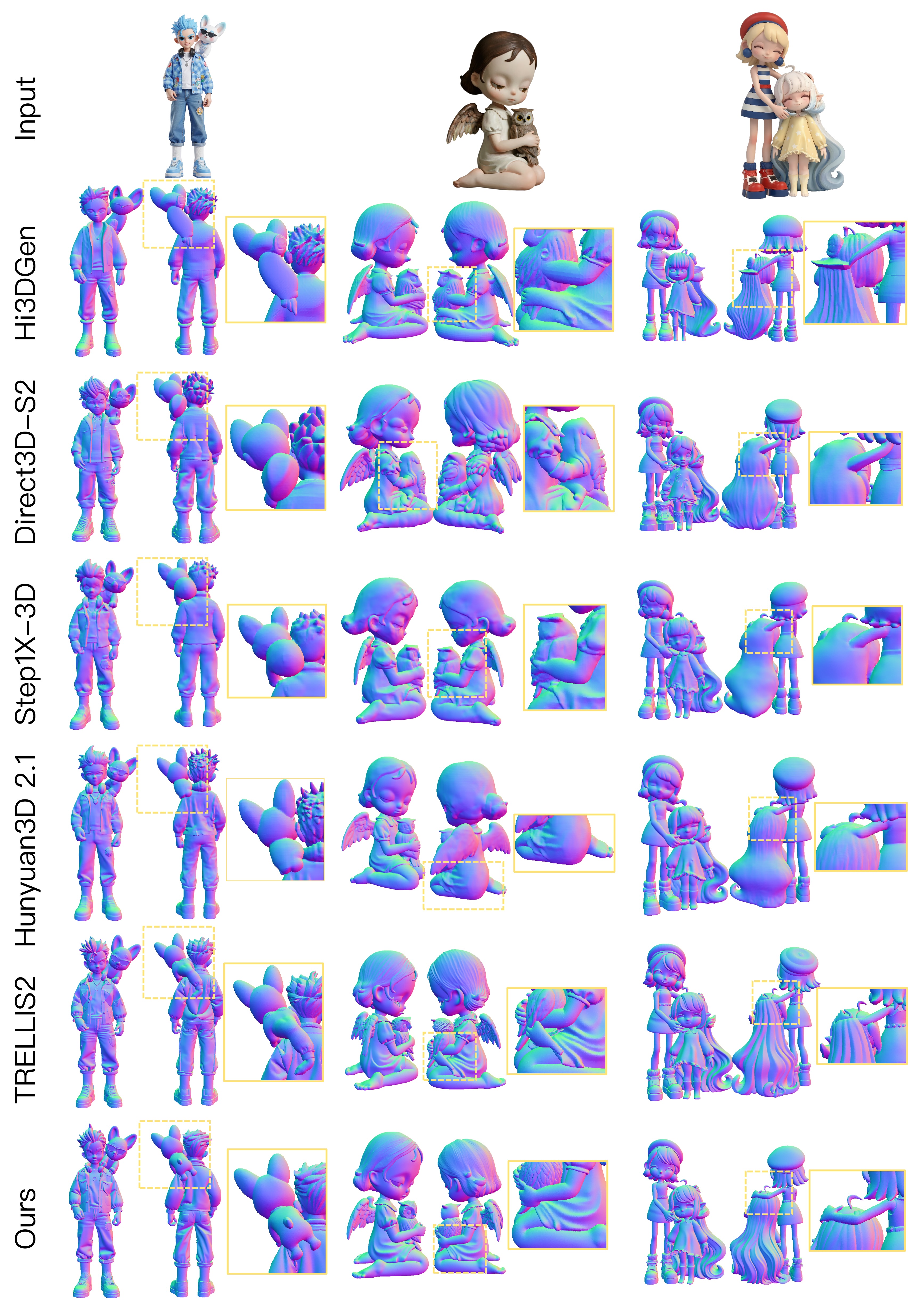} 
  \caption{\textbf{Visual Comparison between Know3D and Current SOTA Baselines.} By leveraging semantic knowledge, our method taps into the intrinsic understanding of object attributes within pre-trained multimodal models, showing potential for improving the structural plausibility of unseen parts in 3D generation.} 
  \label{fig:plausibility}
\end{figure}

\paragraph{Comparison with Single-Image-to-3D Generation Methods.} We conduct comparative experiments with several state-of-the-art, open-source single-image-to-3D generation methods, including Hunyuan3D-2.1~\cite{hunyuan3d2025hunyuan3d}, TRELLIS.2~\cite{xiang2025trellis2}, TRELLIS~\cite{xiang2025trellis}, Step1X-3D~\cite{li2025step1x}, Hi3DGen~\cite{ye2025hi3dgen}, and Direct3D-S2~\cite{wu2025direct3d}. As shown in Tab.~\ref{tab:comparison}, Know3D achieves competitive ULIP and Uni3D scores, indicating effective semantic alignment between the generated meshes and input images. Fig.~\ref{fig:plausibility} presents the qualitative comparison of back-view geometry between Know3D and SOTA baselines, visualized via surface normal maps. Our method leverages semantic knowledge to harness the intrinsic understanding of object attributes within pre-trained multimodal models, demonstrating its potential to enhance the structural plausibility of unseen components in 3D generation. 

\paragraph{Comparison with Hunyuan3D-2mv.} In addition, to analyze the effect of directly using synthesized views as multi-view inputs, we further construct a simple baseline by feeding the input front view together with the generated back-view into a multi-view 3D generation model. In this setting, we adopt Hunyuan3D-2mv~\cite{hunyuan3d22025tencent} as the multi-view baseline for comparison. Despite being trained on large-scale 3D data for 3D generation, Hunyuan3D-2mv is outperformed by Know3D in terms of ULIP and Uni3D scores as shown in Tab.~\ref{tab:comparison}. This validates the effectiveness of our Knowledge Extraction and Prompting design. As illustrated in Fig.~\ref{fig:ab_mv}, although both methods achieve reasonable alignment with the conditioned views (a),(b),  Hunyuan3D-2mv generates distorted and implausible geometries (d), whereas Know3D maintains consistent structures (e).

\subsubsection{Back-View Semantic Controllability.}
Know3D's key advantage over existing methods is its ability to semantically control back-view content via natural language instructions. Existing single-view 3D generation methods can only replicate the visible front view, while back-view generation remains stochastic and cannot respond to user-specified requirements for occluded regions. Fig.~\ref{fig:senamic} demonstrates our model's flexible semantic controllability over the back-view. The leftmost column shows the input front-view images. The second column presents the unconstrained back-view completion results of Know3D, which already produce geometrically plausible and semantically consistent structures aligned with the front view. The subsequent columns present semantically controlled back-view synthesis results guided by different text prompts. These results show that our model can modify the unseen back-side content according to user instructions, while preserving geometric consistency with the original front view. 

\begin{table}[t]
  \centering
  \small
  \setlength{\tabcolsep}{6pt}
  \caption{Quantitative comparison on HY3D-Bench~\cite{hunyuan3d2026hy3d}. 
  The best results are in bold and the second-best are underlined. 
  $\uparrow$ indicates higher is better.}
  \label{tab:comparison}
  \begin{tabular}{lcccc}
    \toprule
    & \multicolumn{2}{c}{HY3D-Bench-test}
    & \multicolumn{2}{c}{HY3D-Bench-val} \\
    \cmidrule(lr){2-3} \cmidrule(lr){4-5}
    Model
    & ULIP $\uparrow$ & Uni3D $\uparrow$
    & ULIP $\uparrow$ & Uni3D $\uparrow$ \\
    \midrule

    Direct3D-S2~\cite{wu2025direct3d}
    & 0.2034 & 0.3292
    & 0.2040 & 0.3303 \\

    Hi3DGen~\cite{ye2025hi3dgen}
    & 0.2082 & 0.3206
    & 0.2030 & 0.3213 \\

    Step1X-3D~\cite{li2025step1x}
    & 0.2068 & 0.3233
    & 0.2033 & 0.3265 \\

    TRELLIS~\cite{xiang2025trellis}
    & \underline{0.2143} & 0.3421
    & \textbf{0.2133} & \underline{0.3464} \\

    TRELLIS.2~\cite{xiang2025trellis2}
    & 0.1948 & 0.3308
    & 0.1967 & 0.3358 \\

    Hunyuan3D-2MV~\cite{hunyuan3d22025tencent}
    & 0.2108 & 0.3413
    & 0.2111 & 0.3400 \\

    Hunyuan3D-2.1~\cite{hunyuan3d2025hunyuan3d}
    & 0.2140 & \underline{0.3434}
    & 0.2122 & 0.3433 \\

    \textbf{Ours}
    & \textbf{0.2174} & \textbf{0.3518}
    & \underline{0.2127} & \textbf{0.3512} \\

    \bottomrule
  \end{tabular}
\end{table}

\subsection{Ablation Study}
We conduct ablation studies to explore two questions in our knowledge prompting design. First, we investigate how the choice of MMDiT timestep for feature extraction influences generation performance—do earlier or later stages of the denoising process provide more useful priors. Second, we compare different feature representations—MMDiT~\cite{wu2025qwen} latent hidden states, VAE encoder~\cite{wan2025} features, and DINOv3~\cite{simeoni2025dinov3} features—to assess what type of information is most effective for guiding 3D generation. We rendered front and back views of a 100-randomly-selected 3D assets subset and extracted their sparse voxels as ground truth for subsequent comparisons. We directly use the ground truth front and back views as the condition to compare the capabilities of different features. For all ablation experiments, we adhered to the same experimental setup: training the first-stage DiT, which generates sparse voxels on 60k training samples with a batch size of 64 for 70k steps. For the original parameters of TRELLIS.2, we employed LoRA (rank=64) for adaptation, while the newly added modules were fully fine-tuned.

\subsubsection{Different Timesteps in MMDiT.}
We evaluate the impact of the denoising timestep $t$ used to extract the MMDiT hidden states, considering $t \in \{0, 0.25, 0.5, 0.75\}$.  As shown in Table~\ref{tab:ablation_timestep}, $t=0.25$ achieves the best overall performance, yielding the highest IoU and lowest Chamfer Distance (CD). We hypothesize that at this intermediate stage ($t=0.25$), the MMDiT has already resolved the global layout and core semantic components of the back-view. Features from earlier timesteps may focus on low-level pixel-wise details that are less aligned with 3D structural priors, whereas later stages are subject to increased noise interference, leading to less reliable guidance for the 3D generation backbone.
\begin{figure}[t] 
  \centering
  \includegraphics[width=1\linewidth]{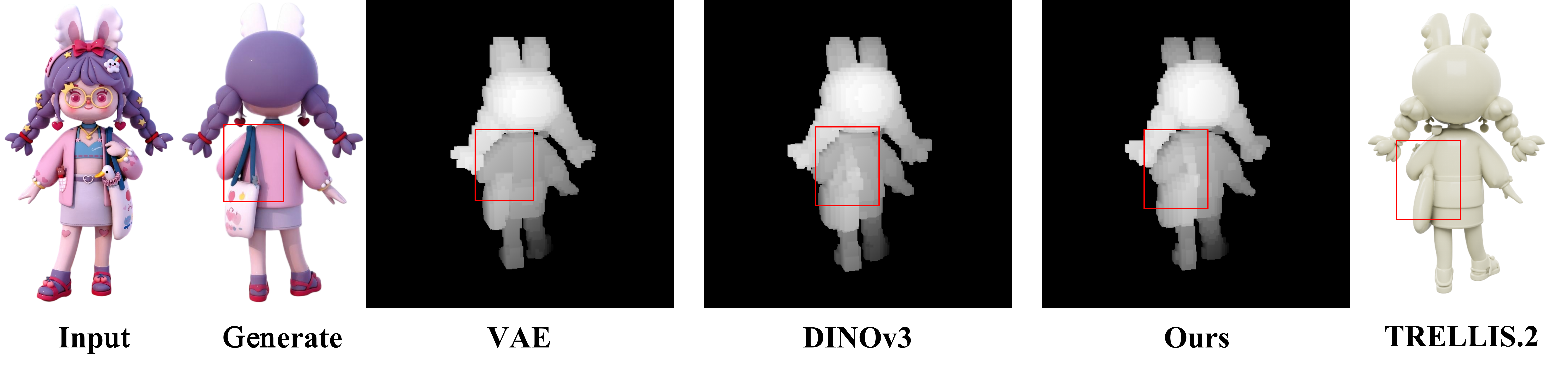} 
  \caption{\textbf{Visualization of Different Feature Representations to Prompt 3D Generation.} When Qwen-Image generates an unreasonable back-view, using different feature representations to prompt 3D generation will lead to different results. The hidden states from the intermediate layers of MMDiT during the denoising process still generate reasonable results, while the VAE fails to inject the back-view information, making it quite similar to the unreasonable results of the original TRELLIS.2. Additionally, when the generated image is directly passed through DINOv3, DINOv3 tends to fit the erroneous results.} 
  \label{fig:ab_feature}
\end{figure}

\begin{table}[!htbp] 
  \centering
  \begin{minipage}[t]{0.48\linewidth}
    \centering
    \caption{Ablation on MMDiT timestep. The best results are in bold and the second-best are underlined. $\uparrow$/$\downarrow$ indicate higher/lower is better.}
    \label{tab:ablation_timestep}
    \begin{tabular}{lcc}
      \toprule
      \textbf{Timestep $t$} & \textbf{IoU $\uparrow$} & \textbf{CD $\downarrow$} \\
      \midrule
      0.0   & 0.343 & 2.376 \\
      0.25  & \textbf{0.352} & \textbf{2.262} \\
      0.5   & \underline{0.349} & \underline{2.272} \\
      0.75  & 0.336 & 2.452 \\
      \bottomrule
    \end{tabular}
  \end{minipage}
  \hfill
  \begin{minipage}[t]{0.48\linewidth}
    \centering
    \caption{Ablation on different feature representations. The best results are in bold and the second-best are underlined. $\uparrow$/$\downarrow$ indicate higher/lower is better.}
    \label{tab:ablation_featuresource}
    \begin{tabular}{lcc}
      \toprule
      \textbf{Feature Source} & \textbf{IoU $\uparrow$} & \textbf{CD $\downarrow$} \\
      \midrule
      VAE encoder      & 0.308 & 2.803 \\
      DINOv3           & \underline{0.342} & \underline{2.385} \\
      MMDiT ($t=0.25$) & \textbf{0.352} & \textbf{2.262} \\
      \bottomrule
    \end{tabular}
  \end{minipage}
\end{table}

\subsubsection{Different Feature Representations to Prompt.}
We further investigate the effectiveness of various feature representations for guiding 3D generation. Specifically, we experiment with (1) directly using the fully denoised VAE~\cite{wan2025} latent from MMDiT, (2) decoding this fully denoised VAE latent into an image and then extracting features via DINOv3~\cite{simeoni2025dinov3}, and (3) directly using the hidden states from the intermediate layers of MMDiT during the denoising process.

As shown in Table~\ref{tab:ablation_featuresource}, the MMDiT hidden states extracted at timestep $t=0.25$ consistently outperform other feature representations on both metrics, achieving the best overall performance. This suggests that MMDiT hidden states contain rich priors of 3D semantics and structure. We have also observed similar phenomena in recent studies~\cite{huang2025much3d}. In contrast, the VAE encoder features yield the lowest performance. This may be because VAE latents are primarily optimized for pixel-level reconstruction, which tends to preserve low-level appearance information while discarding higher-level semantic and structural cues that are important for 3D generation. Moreover, as shown in Fig.~\ref{fig:ab_feature}, when leveraging the pipeline of Qwen-Image to guide 3D generation, inconsistencies may occasionally arise. For instance, the front view might depict a single-shoulder bag, while the generated back-view incorrectly includes two straps. Using different feature representations to prompt 3D generation will lead to different results. The VAE fails to inject the back-view information, making it quite similar to the impossible results of the original TRELLIS.2. Additionally, when the generated image is directly passed through DINOv3, DINOv3 tends to fit the erroneous results. The hidden states from the intermediate layers of MMDiT during the denoising process still generate reasonable results. This shows that it not only has strong 3D perception ability but also contains rich semantic information.

\section{Limitation and Conclusion}
In this paper, we propose Know3D, a novel knowledge-guided 3D generation framework designed to address the inherent ambiguity and lack of semantic control in single-view 3D synthesis. By leveraging a large multimodal model as a ``semantic brain'', we successfully bridge the gap between abstract textual instructions and the geometric reconstruction of the unobserved regions, transforming the traditionally stochastic back-view hallucination into a semantically-controllable process.  We extract rich structural–semantic priors from the intermediate MMDiT layers to prompt the 3D generation model. Our ablation studies further confirm that intermediate denoising states from the multimodal diffusion process capture essential 3D structural and semantic priors.

\textbf{Limitation.} While Know3D  achieves semantic controllability in 3D generation by leveraging multimodal priors, the structural robustness of the generated assets is still influenced by the underlying multimodal foundation models. When the multimodal foundation model cannot fully understand the instructions, then the 3D generation will still be misled to incorrect 3D shapes. 
This could potentially be alleviated by adopting stronger MLLMs or exploring more effective ways of leveraging multimodal guidance and information injection in the 3D generation process.

\section*{Acknowledgements}
This work was financially supported by the Guangdong Provincial Key Laboratory of Ultra High Definition Immersive Media Technology (Grant No. 2024B1212010006), the Outstanding Talents Training Fund in Shenzhen, the Shenzhen Science and Technology Program (Grant Nos. SYSPG20241211173440004 and KJZD20230923114707016), the R24115SG MIGU-PKU META VISION TECHNOLOGY INNOVATION LAB., and the National Natural Science Foundation of China (Grant No. 62272142).
%
%
\bibliographystyle{splncs04}
\bibliography{main}

\end{document}